\documentclass[smallextended]{svjour3}
\usepackage{fix-cm}
\usepackage{bbm}
\usepackage{amsmath}
\usepackage{amssymb}
\usepackage{url}
\usepackage{longtable}
\newcounter{countalg}
\newenvironment{algorithm}[1]%
{\refstepcounter{countalg}%
\begin{minipage}{\textwidth}\baselineskip=1.5mm %
\setlength\unitlength{1truecm}%
\begin{picture}(15,0)(0,0)\put(0,0){\line(1,0){11.882}}\end{picture}\vspace{2mm}\\%
{\bf Algorithm \thecountalg}~#1\\\begin{quote}\vspace{-1mm}~}%
{\end{quote}%
\setlength\unitlength{1truecm}%
\begin{picture}(15,0)(0,0)\put(0,0){\line(1,0){11.882}}\end{picture}%
\end{minipage}}
\newenvironment{routine}%
{\begin{minipage}{\textwidth}%
\setlength\unitlength{1truecm}%
\begin{picture}(15,0)(0,0)\put(0,0){\line(1,0){11.882}}\end{picture}%
\baselineskip=1.5mm\begin{quote}}%
{\end{quote}%
\setlength\unitlength{1truecm}%
\begin{picture}(15,0)(0,0)\put(0,0){\line(1,0){11.882}}\end{picture}%
\end{minipage}}

\journalname{Quantum Information Processing}
\begin{document}
\title{A quantum genetic algorithm with quantum crossover and mutation operations}
\titlerunning{A quantum genetic algorithm with quantum crossover and mutation operations}
\author{Akira SaiToh \and Robabeh Rahimi \and Mikio Nakahara}
\institute{A. SaiToh \at
Quantum Information Science Theory Group, National Institute
of Informatics, 2-1-2 Hitotsubashi, Chiyoda, Tokyo 101-8430, Japan
\and
R. Rahimi \at
Institute for Quantum Computing, University of Waterloo,
200 University Avenue West, Waterloo, Ontario N2L 3G1, Canada
\and
M. Nakahara \at
Department of Physics, Kinki University, 3-4-1 Kowakae,
Higashi-Osaka, Osaka 577-8502, Japan
\and
A. SaiToh \and M. Nakahara \at
Research Center for Quantum Computing, Interdisciplinary
Graduate School of Science and Engineering, Kinki University,
3-4-1 Kowakae, Higashi-Osaka, Osaka 577-8502, Japan
\and
Present Address:\\
A. SaiToh \at
Department of Computer Science and Engineering,
Toyohashi University of Technology,
1-1 Hibarigaoka, Tenpaku-cho, Toyohashi, Aichi 441-8580, Japan\\
\email{saitoh@sqcs.org}
}

\date{Received: date / Accepted: date}

\maketitle

\begin{abstract}
In the context of evolutionary quantum computing in the literal meaning, a quantum
crossover operation has not been introduced so far. Here, we introduce a novel quantum
genetic algorithm which has a quantum crossover procedure performing crossovers among
all chromosomes in parallel for each generation. A complexity analysis shows that a
quadratic speedup is achieved over its classical counterpart in the dominant factor of
the run time to handle each generation.
\end{abstract}



\section{Introduction}\label{secInt}
Continuous development has been performed on genetic algorithms \cite{H75,G89,Mi96}.
Along with the development of quantum computing \cite{G99,NC2000},
quantum-inspired classical algorithms for evolutionary computing
have been developed \cite{NM96,HK00,HK02,HK04,NIOI06,NIO06,CQ07,LWQ10,MEEH12} (see
also a review \cite{Zh10} and references therein).
In addition, classical genetic algorithms to evolve quantum circuits
have also been studied by several authors
\cite{WG98,Sp98,BBS99,Sp99,YI00,Ru01,LP02,LB03,Sp04,LB04,MCS04,Lu04,MCS06,D06,SpK08}
(see also review articles \cite{GPT04,GS07}).
These algorithms are, however, designed to work on classical computers.
Quantum genetic algorithms (QGAs), in its literal meaning, nonetheless,
have gathered comparably little attention and a few works
\cite{RSFA01,UPV04,MBC04,UPV06,MBC08} have been performed so far. Evolutionary
computing on quantum architectures will achieve more attention if there
is a scenario to establish significant improvement over classical
counterparts. Indeed, Malossini {\em et al.} \cite{MBC04,MBC08} claimed
that the computational complexity of an evolutionary step between
generations is exponentially fast in their quantum algorithm in
comparison to a classical one. It seems, however, that they overlooked the
complexity of index-to-string conversion circuits or, otherwise, overlooked
the circuit complexity of a variant of the inversion-about-average operation.
Their claim is thus misleading as we will discuss in Sect.\ \ref{secM}.
Recently, Johannsen {\em et al.}~\cite{Jo10} applied quantum search algorithms to
several optimization problems in a certain context of evolutionary computing. Nevertheless,
it is obscure how much cost is spent for the internal quantum circuit of
a variant of the inversion-about-average operation used for amplitude
amplification (namely, the operation denoted as
$\mathcal{A}{\rm S}_0^\phi\mathcal{A}^{-1}$ in the convention of
Ref.~\cite{BHT98}) in their approach.

Let us briefly summarize the conventional approach of QGAs and its
problem. Here, we omit discussions on incomplete or physically unfeasible
works on QGAs, which were summarized by Sofge in Ref.~\cite{So08}.
The aim of a genetic algorithm is, in most cases, to find an individual
(typically, an input string) with a very high fitness value for a given problem.  
It starts with $O(|A|^w)$ initial individuals where $|A|$ is the size of
the alphabet $A$ and $w$ is the length of a schema expected to be a
building block for a given problem. Let us restrict the problem by representing
each individual as a chromosome encoded as a binary string with the length $n$.
We need $\sim N=2^n$ initial chromosomes for the worst case.
With a quantum register, one may use $n$ qubits to make a superposition of $N$
chromosomes. We regard the probability to find a chromosome on the computational
basis as its (normalized) population.

First we briefly overview the selection strategies in conventional QGAs. In short,
a selection is an operation to enhance populations of individuals with high fitness
values and to decrease those with low fitness values. In quantum computing, it is natural
to utilize variants of Grover's algorithm \cite{Gr96} for this purpose. In fact, a
variant of the Grover search for {\em a priori} unknown number $r$ of solutions (the
Grover-BBHT search) \cite{BBHT98} and that for finding the maximum \cite{DH96,AK99} are
the essential parts of the QGAs developed in Refs.\ \cite{UPV04,MBC04,UPV06,MBC08}. The
query complexity $O(\sqrt{N/r})$ for a variant of the Grover search dominates the total
complexity in the QGA of Udrescu {\em et al.}~\cite{UPV04,UPV06}. For the QGA of Malossini
{\em et al.}~\cite{MBC04,MBC08}, there is a different factor to consume time. As mentioned,
this will be discussed in detail in Sect.~\ref{secM}.

Second we overview the strategies for crossovers and mutations in conventional QGAs. In a
short explanation of these terms, a crossover is an operation to exchange substrings of two
chromosomes and a mutation is an operation to flip certain bits of a chromosome. These are
effective operations to enlarge the search space. So far, quantum crossover operations have not
been developed.
Malossini {\em et al.}~\cite{MBC04,MBC08} used classical crossover and mutation operations;
Udrescu {\em et al.}~\cite{UPV04,UPV06} did not use crossover and mutation operations.
Johannsen {\em et al.}~\cite{Jo10} introduced quantum mutation operations in
an application of quantum amplitude amplification \cite{BHT98} to some optimization problems;
nevertheless, crossovers were not used. It is, in fact, in general difficult to manipulate populations of
quantum states for handling crossovers if we imitate the classical way in a straightforward manner
(See also Ref.~\cite{GPT04} which explained the difficulty in a slightly different manner).
Picking-up two particular individuals and making a crossover costs $O(n)$ quantum gates
if we use $O(n)$ ancillary qubits. Since there are many possible pairs for a crossover at each
generation, there is no speedup over a classical crossover. More specifically speaking, this
approach needs to look up classical data of chromosomes to specify a pair of chromosomes. Thus
there cannot be any speedup. In addition, this requires an exponentially large classical memory
in comparison to the size of a quantum register. We will show a different approach to handle
crossovers in our algorithm in Sect.~\ref{secAlg}.

As another direction for developing quantum evolutionary computing, which we
do not pursue in this contribution, one may use so-called quantum fixed-point search
algorithms \cite{Gr05,TGP06}.
They increase the population of a superposition of target chromosomes by iterative
applications of unitary amplitude-amplification operations. Unlike the
original Grover search, there is no drawback in an excessive iteration;
the probability of finding a target grows monotonically as $1-\epsilon^{2t+1}$
where $\epsilon$ is the probability of finding a non-target chromosome and $t$ is
the number of queries.
The query complexity of this approach is, however, as large as
that of the exhaustive search: When the number $r$ of targets is very small
in comparison to $N$, we have $\epsilon^{2t+1}=(1-r/N)^{2t+1} \simeq 1-(2t+1)r/N$,
which implies that $O(N/r)$ queries are required to achieve a sufficiently small
error probability, say $1/2$. Thus for $r\ll N$, Grover's algorithm
should be chosen instead of fixed-point quantum search algorithms.
In addition, the asymptotic optimality of $\epsilon^{2t+1}$ as an error
reduction speed was proved \cite{CRR05} for fixed-point quantum search
methods. It is also known that the query complexity of the Grover search
is optimal \cite{BBHT98} in general as a unitary process for unsorted search.
Therefore, it is unlikely\footnote{{
There is another drawback in the use of a quantum fixed-point search. It requires
phase shift operations
$\widetilde{U}_{\rm s}=1-(1-e^{i\pi/3})\sum_l|\varsigma_l\rangle\langle \varsigma_l|$
and $\widetilde{U}_{\rm t}=1-(1-e^{i\pi/3})|\tau_m\rangle\langle \tau_m|$ with
{\em source states} $|\varsigma_l\rangle$ and {\em target states} $|\tau_m\rangle$ ($l$ and $m$
are labels) in addition to another appropriate unitary operation \cite{Gr05}.
One may alternatively use ${\hat{U}_{\rm s}}=1-(1-e^{i\pi/3})|\mathcal{S}\rangle\langle \mathcal{S}|$
and ${\hat{U}_{\rm t}}=1-(1-e^{i\pi/3})|\mathcal{T}\rangle\langle \mathcal{T}|$ where
$|\mathcal{S}\rangle$ is an equally-weighted superposition of $|\varsigma_l\rangle$ and
$|\mathcal{T}\rangle$ is an equally-weighted superposition of $|\tau_m\rangle$.
(This kind of alternative operation for a quantum search was used in Ref.~\cite{T11}.)
The problem is that $\varsigma_l$'s are highly-random nonconsecutive chromosomes in
the context of evolutionary computing. There is no known way to construct
$\widetilde{U}_{\rm s}$ or $\hat{U}_{\rm s}$ within ${\rm poly}(\log \widetilde{N})$
cost in such a case, where $\widetilde{N}$ is the number of $\varsigma_l$'s.
}} that a fixed-point quantum search is effectively used in an evolutionary
computing instead of the standard Grover search and its variants.

In this contribution, we propose a quantum genetic
algorithm that involves a quantum crossover and a quantum mutation. It uses
the quantum search for finding the maximum \cite{DH96,AK99} for the
selection procedure. Although this selection strategy looks similar to
those of conventional QGAs \cite{UPV04,MBC04,UPV06,MBC08}, it is different
in the point that we use tailored inversion operations for the quantum
search. Our algorithm achieves a quadratic speedup over its classical
counterpart in the dominant factor of the run time to update each generation
as an algorithm involving all possible crossovers for a chosen crossing site.

This paper is organized in the following way. Section \ref{secM}
discusses on the algorithm of Malossini {\em et al.} \cite{MBC04,MBC08}.
Section \ref{secAlg} describes our algorithm.
The procedures of the crossover, the mutation and the selection are
developed in the standard quantum circuit model.
Computational complexities of the algorithm are evaluated in Sect.~\ref{seccost}.
The obtained results are discussed in Sect.~\ref{secDis}
and summarized in Sect.~\ref{secCon}.

\section{A conventional quantum genetic algorithm}\label{secM}
Here we discuss the conventional quantum genetic algorithm proposed by 
Malossini {\em et al.} \cite{MBC04,MBC08}. There is another conventional
algorithm proposed by Udrescu {\em et al.} \cite{UPV04,UPV06},
which is not introduced here. The main difference is that Udrescu {\em et
al.} did not employ crossovers and mutations while Malossini {\em et al.}
employed them as classical operations.

The problem we consider is given as follows.\\
{\em Problem:} Suppose there are $N$ individuals with indices
$0,\ldots, N-1$. There is a fitness function $f:\{0,\ldots,N-1\}\rightarrow
[0,1]$. Find one of the individuals with sufficiently large fitness values.\\~

The algorithm of Malossini {\em et al.} is described in Fig.~\ref{figalgM} (Algorithm \ref{algM}).
\begin{figure*}[h]
\begin{algorithm}{(Malossini {\em et al.} \cite{MBC04,MBC08})}\label{algM}
\hspace{-3mm}{\it Main Procedure:}\\
Start from a generation consisting of $\widetilde{N} \le N$ individuals.\\
Repeat:
\begin{itemize}
\item[1.] Use the {\it Quantum Selection Procedure} to get one index.
Do the same procedure to get another.
\item[2.] Classically make a crossover for the two individuals
corresponding to the indices obtained in 1. Classically make a proper mutation.
\item[3.] Classically replace randomly chosen two individuals with the
two offsprings obtained in 2.
\end{itemize}
~\\
{\it Quantum Selection Procedure:}
\begin{itemize}
\item[(i)] Choose randomly an index $y$ among $\widetilde{N}$ possible ones in the present
generation and compute its fitness $f(y)$.
Set the threshold $F_y\longleftarrow f(y)$.
\item[(ii)] Perform $\tau$ times:
\begin{itemize}
\item[(a)] Initialize the quantum registers to
$\frac{1}{\sqrt{\widetilde{N}}}\sum_{x}|x\rangle|y\rangle$, where $x$'s are the indices of
the present generation.
\item[(b)] Apply the Grover-BBHT search \cite{BBHT98}, where we internally
 use an oracle that inverts the signs of the marked states, namely
 states $|x\rangle$ such that $f(x)\ge F_y$. This enhances the
 amplitudes of marked states after its Grover iteration.
\item[(c)] Measure the left register in the computational basis and get the new
 index $y'$. If $f(y')>F_y$, then $y\leftarrow y'$ and $F_y\leftarrow f(y')$.
\end{itemize}
\item[(iii)] Return the index $y$.
\end{itemize}
\end{algorithm}
\caption{\label{figalgM}Description of Malossini {\em et al.}'s algorithm.}
\end{figure*}
They analyzed their algorithm and claimed \cite{MBC04,MBC08} that the selection
procedure internally requires a very small number of oracle calls in contrast to
$O(\widetilde{N}\log \widetilde{N})$ oracle calls required by a classical
selection procedure, where $\widetilde{N}$ is the number of individuals in a
generation. In short, their claim was that the number of the Grover iterations
in step (b) (of Fig.~\ref{figalgM}) was $O(1)$.

Their complexity analysis was, however, misleading. A more accurate
description of the complexity is given in the following way. There are two
possible cases we should consider: (A) $\widetilde{N}$ is approximately equal
to $N$; (B) $\widetilde{N} \lnapprox N$.

Let us consider the case (A). We can preset the index $x$ to represent
the $x$th individual in a good approximation thanks to $\widetilde{N}\simeq N$.
Then the given fitness function can be used as it is, without index conversion.
This indicates that we use nothing but the standard Grover-BBHT search for $N$ indices.
As $\widetilde{N}\simeq N$, the quantum selection procedure is called only once or a
very small number of times. Suppose there are $r$ nearly-optimal individuals that are
acceptable as solutions for a given problem. Inside the quantum selection procedure,
as the selection goes on in (ii), the oracle used in step (b) has
less marked states. The number of marked states converges to
$r$. Therefore, the query complexity of a single step in a very later
selection stage is $O(\sqrt{\widetilde{N}/r})=O(\sqrt{N/r})$. This is the
accurate description of the query complexity for the present case and it is the same
as that Udrescu {\em et al.} gave for their algorithm \cite{UPV04,UPV06}.
On average, each time of the repetition in (ii) virtually extinguishes a half
of the individuals by increasing the threshold. Thus, by considering the sum of
the geometric series $\sqrt{N/(2^ir)}$ ($i=0,1,2,\ldots$), the
average total query complexity is found to be $O(\sqrt{N/r})$.

Let us now consider the case (B). In this case, the quantum selection
procedure is called several (or more) times.
The important fact is that the chromosomes of individuals in a
generation are not consecutive binary strings. The generation
contains $\widetilde{N}$ binary strings out of $N$ possible ones.
There are two options to handle them: (B-1) consecutive integers are
assigned as pointers to nonconsecutive chromosomes; (B-2)
nonconsecutive chromosomes are used as they are.\\
(B-1): In this case, the index $x$ is a pointer to the chromosome $\kappa_x$
for which the fitness function works appropriately.
(In short, this is a workaround for the fact that the original Grover-BBHT
search should start from a uniform superposition of consecutive indices
while the generation consists of nonconsecutive chromosomes.)
Therefore, for the superposition
$|\psi_0\rangle=\frac{1}{\sqrt{\widetilde{N}}}\sum_x|x\rangle|y\rangle$, the
fitness function cannot be applied directly. We need a conversion
circuit to interpret $x$ as $\kappa_x$. Then the fitness function is applied
and it returns $f(\kappa_x)$. Each conversion should involve $O(\log N)$ CCNOT
gates. Thus, the total circuit depth for the index conversions is
$O(\widetilde{N}\log N)$. As a result, a single oracle call accompanies the additional
time complexity $O(\widetilde{N}\log N)$. Indeed, the query complexity is
small if $\widetilde{N}$ is much smaller than $N$. However, the cost of index
conversions hinders speedup over a classical selection procedure.\\
(B-2): In this case, the state $|x\rangle$ in the superposition
$|\psi_0\rangle=\frac{1}{\sqrt{\widetilde{N}}}\sum_x|x\rangle|y\rangle$ keeps the
actual chromosome $x$ as an index. One should use the Grover-search routine somehow
without index conversions by choosing $|\psi_0\rangle$ as its initial state.
Indeed, this is possible \cite{BHT98} in the case where one can provide the
operation $L=I-2|\psi_0\rangle\langle\psi_0|$ as the inversion-about-average
operation instead of the standard one for the original Grover search. Nevertheless,
$L$ should be generated by sandwiching $I-2|0\rangle\langle 0|$ with $U_{\rm (a)}$
and $U_{\rm (a)}^\dagger$ in the present context, where the unitary operation
$U_{\rm (a)}$ corresponds to the initialization step (a). Thus the circuit
complexity of $L$ is $O(\widetilde{N}\log N)$ (see, {\em e.g.}, Ref.~\cite{VM99}).
(There is a little confusing result by Soklakov and Schack \cite{SoSc2006}; they showed that
a state preparation, namely an initialization, is possibly performed within polynomial
cost in $\log N$ for some cases. Nevertheless, considering the internal cost of the special
oracle they use, their method spends $O(\widetilde{N}{\rm poly}(\log N))$ time for general
cases including the present case where the parent set consists of random indices.)
As a consequence, an expensive circuit for $L$ should be used subsequent to every query.
Obviously, the time complexity in this case is as large as the one in case (B-1).

One may also think of replacing the Grover-search routine with the
generalized Grover search for nonconsecutive integer sets \cite{G98}.
Nevertheless, it is required to find an appropriate unitary transformation
replacing the standard Hadamard transformation $H\otimes\cdots\otimes H$.
There is no known way to find it efficiently when a random integer set is
given and target integers are unknown in advance.

There is, in fact, a way to reduce the circuit complexity of $L$ introduced in
(B-2) if the algorithm is modified so that it uses an efficient pseudo-randomizer
(or, pseudo-scrambler) circuit instead of directly using a random number
generator. Our algorithm introduced in the next section takes this approach.

In addition to the above discussions, we should mention that Algorithm~\ref{algM} has
another problem: it takes $O(\widetilde{N}\log N)$ space to keep a generation
in a classical memory. This is usually quite larger than the $O(\log N)$
space that is enough for quantum search (neglecting the space internally used by an
oracle circuit). Our algorithm is designed not to face this problem.

\section{Algorithm with quantum crossover and quantum mutation}\label{secAlg}
As we have seen in the previous sections, conventional quantum genetic algorithms were not
designed to achieve quantum speedup in their selection procedures in case a generation
consists of nonconsecutive chromosomes. In addition, a quantum crossover operation has
not been developed so far.
Here, we propose an algorithm using a quantum crossover and a quantum mutation. We use
a variant of the Grover search with tailored subroutines for the selection procedure,
whose query complexity is the main factor of the total time complexity. We achieve a
quadratic speedup over a classical counterpart in a dominant factor of time complexity
as a genetic algorithm with crossovers among all parents. It is novel in the sense that
a simultaneous crossover using a superposition is achieved. The speedup partly relies on
an efficient internal structure of a pseudo-randomizer circuit, which will be explained in
Appendix~A.

\subsection{Algorithm flow}
We introduce our algorithm in Fig.~\ref{figouralg} (Algorithm \ref{ouralg}). In this algorithm, the
quantum register is accessible from its subroutines as a kind of global variables. The procedures
called inside the algorithm are described in corresponding subsections \ref{secinit}-\ref{secselection}.
\begin{figure*}[h]
\begin{algorithm}{}\label{ouralg}
\hspace{-2.5mm}Consider a threshold $f_{\rm th}$ for fitness values, which is considered to be
sufficiently large.\\ $t\longleftarrow 0$.\\
{\bf REPEAT} 1.-9.:
\begin{itemize}
\item[1.] Construct a pseudo-randomizer $R$ that maps a $c$-bit string to a pseudo-random
          $n$-bit string. It should be implemented as a circuit whose input and output are integer couples
          $(a,0)$ and $(a, R(a))$, respectively, where $a$ is a $c$-bit integer. The circuit should
          consist of ${\rm poly}(cn)$ elementary reversible logic gates.
          We require $R$ to regard $0_{0}\cdots 0_{c-1}$ as an exception
          and map it to $0_0\cdots 0_{n-1}$. We also require $\widetilde{N}=2^c\lnapprox N=2^n$.
          Once constructed, $R$ is fixed until next $t$.
          An explicit example to construct $R$ is given in Appendix~A.
\item[2.] {\bf IF} $t=0$ {\bf THEN} generate a random $c$-bit string $\gamma$ and set $z\longleftarrow R(\gamma)$
          (otherwise, $z$ is the best chromosome found in the $(t-1)$th trial) {\bf ENDIF}
\item[3.] Generate a random $c$-bit string $\gamma'$ and set $u\longleftarrow R(\gamma')$.
\item[4.] {\bf CALL} \verb|init_reg|($R$, $z$) defined in Sect.~\ref{secinit} twice to
          make two identical quantum states that are both
	  $|\varphi\rangle=\frac{1}{\sqrt{\widetilde{N}}}\sum_{x\in X}|a_x\rangle_{\rm a}|x\rangle$
	  where $a_x$ is the address pointing to $x\in X$; $X=\{R(q)\}_q\cup\{z\}$ with
          $q=0_{0}\cdots 0_{c-2}1_{c-1},\ldots,1_{0}\cdots 1_{c-1}$; subscript ``a'' stands for the address portion.
          We write the entire unitary operation of this procedure as $U_{\rm init}$.
          That is, $|\varphi\rangle^{\otimes 2}=U_{\rm init}(|0\rangle_{\rm a}|0\rangle)^{\otimes 2}$.
\item[5.] {\bf CALL} \verb|quantum_crossover|($l$) defined in Sect.~\ref{seccrossover}
	  for the current quantum register, which is a $1$-point crossover with the
          crossing site, chosen at one's convenience, placed between the $(l-1)$th and the $l$th
          bits of a chromosome. All possible crossovers for this crossing site are performed
          simultaneously. This procedure is an identity map as a quantum operation acting
          on $|\varphi\rangle^{\otimes 2}$.
\item[6.] Apply the quantum mutation (Sect.~\ref{secmutation}) to the current quantum register.
          We write the entire unitary operation of this procedure as $U_{\rm mut}$.
          Apply the same mutation classically to $u$.
\item[7.] {\bf CALL} \verb|quantum_selection|($U_{\rm init}$, $U_{\rm mut}$, $u$)
          defined in Sect.~\ref{secselection} for the current quantum register and obtain the
          output chromosome $z$.
\item[8.] {\bf IF} $f(z)\ge f_{\rm th}$ {\bf THEN} {\bf RETURN} $z$ and {\bf EXIT}~~{\bf ENDIF}
\item[9.] $t\longleftarrow t+1$. Refresh the quantum register.
\end{itemize}
\end{algorithm}
\caption{\label{figouralg}Description of our algorithm.}
\end{figure*}
\subsection{Preparing the initial state of a quantum register}\label{secinit}
In this subsection, we define the procedure \verb|init_reg|($R$, $z$).

This procedure is intended to prepare a superposition corresponding to the
generation given as a set $X=\{R(q)\}_q\cup\{z\}$ with
$q=0_{0}\cdots 0_{c-2}1_{c-1},\ldots,1_{0}\cdots 1_{c-1}$
(thus $\sharp X=\widetilde{N} = 2^c$). The procedure starts with the quantum state
$|0\rangle^{\otimes c}_{\rm a}|0\rangle^{\otimes n}$.
The desired superposition is $|\varphi\rangle
=\frac{1}{\sqrt{\widetilde{N}}}\sum_{x\in X}|a_x\rangle_{\rm a}|x\rangle$ with $a_x$
the address pointing $x$. We opt to use consecutive addresses $0,\ldots,2^c-1$.
The procedure is now defined in Fig.~\ref{fig_init_reg}.
The order of time complexity of this procedure equals to the internal circuit
complexity ${\rm poly}(cn)$ of the pseudo-randomizer $R$.
An explicit example to construct $R$ as a quantum circuit is given in Appendix~A.
\begin{figure*}[h]
\begin{routine}
{\bf PROCEDURE} \verb|init_reg|($R$, $z$):
\begin{itemize}
\item[(i)] Make a superposition $\frac{1}{\sqrt{\sharp X}}
\sum_{j=0}^{2^c-1}|j\rangle_{\rm a}|0\cdots 0\rangle$
by applying $H^{\otimes c}\otimes I^{\otimes n}$ to $|0\rangle^{\otimes c}_{\rm a}|0\rangle^{\otimes n}$.
\item[(ii)] Apply the pseudo-randomizer $R$ implemented as a unitary
operation mapping $|j\rangle_{\rm a}|0\rangle$ to $|j\rangle_{\rm a}|x_j\rangle$
with $x_j$ an $n$-bit pseudo-random number for $j=1,\ldots,2^c-1$.\\
By assumption, $|0\rangle_{\rm a}|0\rangle$ is mapped to $|0\rangle_{\rm a}|0\rangle$.
\item[(iii)] Apply a $0_0\cdots0_{c-1}$-controlled $X^{z_0}\otimes\cdots\otimes X^{z_{n-1}}$
to map $|0\rangle_{\rm a}|0\rangle$ to $|0\rangle_{\rm a}|z\rangle$, where
$z_k$ is the $k$th bit of $z$ ($k=0,\ldots,n-1$).
\item[(iv)] {\bf RETURN} the current state, namely $|\varphi\rangle$.
\end{itemize}
\end{routine}
\caption{\label{fig_init_reg}Description of procedure {\tt init\_reg}($R$, $z$).}
\end{figure*}

\subsection{Crossover}\label{seccrossover}
Here, we construct a quantum crossover procedure \verb|quantum_crossover|($l$).
It is a $1$-point crossover acting on the chromosomes simultaneously.

Recall that the original generation is given as a set $X$ of $n$-bit integers $x$
with $\sharp X=\widetilde{N}$.
Each $x$ has its left side $x^{\rm left}$ and its right side $x^{\rm right}$
separated by the crossing site. The crossing site is placed between the $(l-1)$th
and the $l$th qubits as specified by the parameter. Hence, the bit length of
$x^{\rm left}$ is $l$. With the procedure, we generate a superposition
of all the children that are combinations of $x^{\rm left}$'s and
$x^{\rm right}$'s together with their parents with the same weight as children.

The state of the quantum register in the beginning of this procedure is
\[|\varphi\rangle^{\otimes 2}
=\left(\frac{1}{\sqrt{\sharp  X}}\sum_{x\in X} |a_x\rangle_{\rm a}|x^{\rm left}\rangle|x^{\rm
 right}\rangle\right) \otimes
 \left(\frac{1}{\sqrt{\sharp  X}}\sum_{x'\in X} |a_{x'}\rangle_{\rm a}|{x'}^{\rm left}\rangle|{x'}^{\rm
 right}\rangle\right).
\]
This state has the components $|x^{\rm left}\rangle|x^{\rm right}\rangle
|{x'}^{\rm left}\rangle|{x'}^{\rm right}\rangle$ besides the addresses.
We relabel the qubits so that the middle portion
$|x^{\rm right}\rangle |{x'}^{\rm left}\rangle$ is put aside from
our minds. Let us conceal them by denoting as $|*_{xx'}\rangle$.
In addition, we denote the main portion $|x^{\rm left}{x'}^{\rm right}\rangle$
with the subscript ``${\rm main}$''.
The state $|\varphi\rangle^{\otimes 2}$ with the new qubit labels is written as
\[
 \frac{1}{\sharp X}\sum_{x\in X}\sum_{x'\in X}
|a_xa_{x'}\rangle_{\rm a}|x^{\rm left}{x'}^{\rm right}\rangle_{\rm main}|*_{xx'}\rangle.
\]
We have at most $(\sharp X)^2$ distinct chromosomes in this state.
In this way, all possible crossovers are performed at once by the relabelling.
The resultant state is a superposition of all of the children together with
their parents. (The parents are involved because the values of $x$ and
$x'$ may coincide.) This is desirable as a crossover because sometimes
some parents have higher fitness values than any child.

The procedure described above is formally written as shown in Fig.~\ref{fig_quantum_crossover}.
\begin{figure*}[h]
\begin{routine}
{\bf PROCEDURE} \verb|quantum_crossover|($l$):
\begin{itemize}
\item[(i)] We have the quantum register in the state $|\varphi\rangle^{\otimes 2}$.
    The original labels of its qubits are $0,\ldots,2c+2n-1$.
    We relabel them as $\underbrace{0,\ldots,c-1}_{c}$, $\underbrace{2c,\ldots,2c+l-1}_{l}$,
$\underbrace{2c+n,\ldots,2c+2n-l-1}_{n-l}$, $\underbrace{c,\ldots,2c-1}_{c}$,
$\underbrace{2c+2n-l,\ldots,2c+2n-1}_{l}$, $\underbrace{2c+l,\ldots,2c+n-1}_{n-l}$.
\item[(ii)] {\bf RETURN}
\end{itemize}
\end{routine}
\caption{\label{fig_quantum_crossover}Description of procedure {\tt quantum\_crossover}($l$).}
\end{figure*}

As is obvious, this procedure is an identity map as a quantum operation.
Once the crossover process is completed, one may use a mutation as an option.
This is going to be explained in the next subsection.

\subsection{Mutation}\label{secmutation}
In classical genetic algorithms, randomly selected chromosomes are affected by a mutation,
which is typically certain bit flips acting on randomly-chosen places.
Here, we consider the mutation procedure described in Fig.~\ref{fig_tmp_mut}.
Although it is written as a classical routine, it can be trivially interpreted as
a quantum circuit.
\begin{figure*}[h]
\begin{routine}
{\bf PROCEDURE} \verb|tmp_mut|:
\begin{itemize}
\item[(i)]
Let us randomly generate the first template like
\verb|***0*1***1**0****1*|, which specifies a schema to mutate. Using this
template, we pick up chromosomes with specified bits like 0,1,1,0,1 in the
specified places.
\item[(ii)] We also use the second template like
\verb|*X*************X***| in which \verb|X|'s can be placed only on
the places where \verb|*|'s (namely, ``don't care'' symbols) are placed in
the first template. Using this template, we apply the bit flip \verb|X| to the
specified places of the chromosomes picked up in (i).
\item[(iii)] {\bf RETURN}
\item[Note:]Technically, we often wish to avoid a mutation for the best chromosome $z$
found so far by the present time step. This is realized by choosing the first template
so that this does not happen.
\end{itemize}
\end{routine}
\caption{\label{fig_tmp_mut}Description of procedure {\tt tmp\_mut}.}
\end{figure*}
As a quantum circuit, this mutation procedure \verb|tmp_mut| is realized by a
multiple-bit controlled multiple-bit NOT gate. (In the example mentioned in
Fig.~\ref{fig_tmp_mut}, the gate is ``{\tt 0-controlled 1-controlled 1-controlled 0-controlled 1-controlled
NOT NOT}'' with control bits specified by the first template and the
target bits specified by the second template.) The gate acts on the portion
$|x^{\rm left}{x'}^{\rm right}\rangle_{\rm main}$; the addresses $a_x$ and
$a_{x'}$ are untouched. Therefore, the resultant state can be written as
\begin{equation}\label{aftermutation}
 |\widetilde{\varphi}\rangle=
\frac{1}{\sharp X}\sum_{x\in X}\sum_{x'\in X}
|a_xa_{x'}\rangle_{\rm a}|\widetilde{x}^{\rm left}\widetilde{x'}^{\rm right}\rangle_{\rm main}|*_{xx'}\rangle,
\end{equation}
where $(\widetilde{x}^{\rm left}\widetilde{x'}^{\rm right})$'s are
the chromosomes after the mutation process.

The next step is to apply a natural selection to the chromosomes
living in the superposition $|\widetilde{\varphi}\rangle$.

\subsection{Selection}\label{secselection}
In this subsection, we introduce our selection procedure.
It is intended to find a chromosome having the maximum fitness
among $(\widetilde{x}^{\rm left}\widetilde{x'}^{\rm right})$'s.
It utilizes the quantum search for finding the maximum \cite{DH96,AK99}. As we have
mentioned, conventional QGAs \cite{UPV04,MBC04,UPV06,MBC08} have similar selection
strategies. The difference from them is that we use tailored inversion operations
for the quantum search.

Our selection procedure is called with three arguments: $U_{\rm init}$, $U_{\rm mut}$ and $u$.
We have the state (\ref{aftermutation}) at the beginning of this procedure.
The procedure is now defined in Fig.~\ref{fig_quantum_selection}. Here in the text, we do not
repeat its description. It should be noted that, in the procedure, we set
$k_{\rm term}=\eta\times\lceil(45/2)\widetilde{N} +(28/5)(\log_2 \widetilde{N})^2\rceil$ with
integer constant $\eta\ge 1$.
\begin{figure*}[htb]
\begin{routine}
{\bf PROCEDURE} \verb|quantum_selection|($U_{\rm init}$, $U_{\rm mut}$, $u$):\\
{\bf FOR} $k\longleftarrow 0$ {\bf TO} $k_{\rm term}-1$:
\begin{itemize}
\item[(i)] 
Set \[
U_1 = I\otimes I\otimes I-2\sum_{f(y)\ge f(u)}I\otimes|y\rangle\langle y|\otimes I
\]
where the left and the right $I$'s act on the address states $|a_xa_{x'}\rangle$ and
the states $|*_{xx'}\rangle$, respectively. This is the oracle function that can be implemented as follows.
First, we attach ancillary qubits as blocks ${\rm (I)}$ and ${\rm (II)}$ in the state
$|0\cdots0\rangle_{\rm (I)}|-\rangle_{\rm (II)}$ with $|-\rangle=(|0\rangle-|1\rangle)/\sqrt{2}$.
For each $xx'$, $f(\widetilde{x}^{\rm left}\widetilde{x'}^{\rm right})$ is set as a value of
the block ${\rm (I)}$ by using $f$ implemented as a quantum circuit. Let us write this operation as $U_f$.
Then we compare the block ${\rm (I)}$ with $f(u)$; we flip the qubit ${\rm (II)}$ if
$f(\widetilde{x}^{\rm left}\widetilde{x'}^{\rm right})\ge f(u)$.
We apply ${U_f}^\dagger$ (this disentangles the ancillary qubits from the main register) and
remove the ancillary qubits.
\item[(ii)]
Set
\[
U_2 = I\otimes I\otimes I
- 2|\widetilde{\varphi}\rangle\langle\widetilde{\varphi}|.
\]
This operation is composed in the following way.
\[
U_2=U_{\rm mut}U_{\rm init}
[I-2(|0\rangle^{\otimes c}|0\rangle^{\otimes n}
\langle 0|^{\otimes c}\langle 0|^{\otimes n})^{\otimes 2}]
U_{\rm init}^\dagger U_{\rm mut}^\dagger,
\]
where we also relabel qubits according to \verb|quantum_crossover|($l$).
\item[(iii)]
Apply the Grover-BBHT search \cite{BBHT98} for which we use $U_1$ and
$U_2$ instead of the standard operations, namely, the inversion operation
for targets and the inversion-about-average operation, respectively.
As for the starting state of the search, we use $|\widetilde{\varphi}\rangle$.
\item[(iv)]
Measure the main register and obtain the chromosome $u'$.\\
{\bf IF} $f(u')>f(u)$ {\bf THEN} $u\longleftarrow u'$ {\bf ENDIF}
\end{itemize}
{\bf END FOR}\\
{\bf RETURN} $u$.
\end{routine}
\caption{\label{fig_quantum_selection}Description of procedure
{\tt quantum\_selection}($U_{\rm init}$, $U_{\rm mut}$, $u$).}
\end{figure*}

Let us give an explanation about $k_{\rm term}$, namely, the number of iterations.
The defined procedure is the same as the well-known quantum search algorithm
for finding the maximum \cite{DH96,AK99} except for the definitions
of the inversion operation for targets and the inversion-about-average
operation. 
In other words, we perform the quantum search for finding the maximum
in the subspace ~${\rm span}\{|a_xa_{x'}\rangle_{\rm a}|\widetilde{x}^{\rm left}
\widetilde{x'}^{\rm right}\rangle_{\rm main}|*_{xx'}\rangle\}.$
As proved by D\"{u}rr and H{\o}yer \cite{DH96},
the probability for the output to be the maximum is at least $1/2$
if the number of iterations is $\lceil (45/2)\sqrt{M}+(7/5)(\log_2 M)^2\rceil$
with $M$ the number of indices. In the present context, we have
$M={\widetilde{N}}^2$ since there are ${\widetilde{N}}^2$ distinct addresses.
Therefore, after $k_{\rm term}$ iterations, we find the output chromosome
having the maximum fitness among $(\widetilde{x}^{\rm left}\widetilde{x'}^{\rm right})$'s,
with the probability at least $1-(1/2)^\eta$.
By setting $\eta$ sufficiently large, say, around $16$ to $24$,
we have the desired output with almost certainty.

\section{Computational cost}\label{seccost}
We are going to evaluate the computational cost of each process of Algorithm \ref{ouralg}
(Fig.~\ref{figouralg}) in Sect.~\ref{seceachcost}.
The total computational cost to handle a generation will be derived in Sect.~\ref{sectotalcost}
and compared with that of classical counterpart in Sect.~\ref{seccomparison}.
\subsection{Costs of each procedure}\label{seceachcost}
For a single call of each procedure, the costs are evaluated as follows.
\paragraph{Cost of the initialization}
The procedure to prepare the initial state, described in Sect.~\ref{secinit},
uses ${\rm poly}(cn)={\rm poly}(\log\widetilde{N}\log N)$ elementary quantum gates.
\paragraph{Cost of the crossover}
The crossover described in Sect.~\ref{seccrossover} makes use of two identical
$(c+n)$-qubit states. This procedure does not use any quantum operation but relabels
qubits. This takes $O(\log N)$ time.
\paragraph{Cost of the mutation}
The mutation described in Sect.~\ref{secmutation} involves a single gate that looks
like, say,
\[\begin{split}
&{\rm \mathrm{*}-NOT-\mathrm{*}-C_0-\mathrm{*}-C_1-\mathrm{*}-\mathrm{*}-\mathrm{*}-
C_1-\mathrm{*}-\mathrm{*}}\\
&{\rm-C_0-\mathrm{*}-\mathrm{*}-NOT-\mathrm{*}-C_1-\mathrm{*}}
\end{split}\]
with multiple control bits (0-control ${\rm C_0}$'s and 1-control ${\rm C_1}$'s) and multiple NOT
gates placed according to the corresponding templates, where symbol $\mathrm{*}$ stands for an
untouched qubit. Such a gate can be realized by $O(\log N)$ elementary quantum gates with $O(\log N)$
ancillary qubits.
\paragraph{Cost of the selection}
The selection described in Sect.~\ref{secselection}
consumes $O(\widetilde{N})$ queries to find the best chromosome in the
set of at most ${\widetilde{N}}^2$ chromosomes. Each query accompanies the operation
$U_2$ that invokes $U_{\rm init}$ and $U_{\rm mut}$, and also their inverse operations.
It is easy to find that the internal cost of $U_2$ is ${\rm poly}(cn)$ according to the
costs for $U_{\rm init}$ and $U_{\rm mut}$.
In addition, the cost to prepare the starting state is ${\rm poly}(cn)$.
We may also mention that, the internal cost of the fitness function is a certain
small factor, typically ${\rm poly}(n)$, as conventionally assumed \cite{G89}.
Therefore, the circuit complexity of the procedure is
$O({\widetilde{N}}{\rm poly}(\log\widetilde{N}\log N))$. (We assume that the circuit depth
is on the order of the circuit size, namely, the number of elementary quantum gates.)
As for space, we use ${\rm poly}(\log N)$ qubits in total, considering a typical fitness
function consuming ${\rm poly}(n)$ space. When the fitness function is designed to use
$O(n)$ space, $O(\log N)$ qubits are enough, although we do not assume this case for
evaluating the space complexity.
\subsection{Total cost}\label{sectotalcost}
Comparing the costs of the four procedures, the dominant cost is
the circuit complexity for the selection procedure. Therefore, we find that
our algorithm uses
\begin{equation}\label{cxtime}
O({\widetilde{N}}{\rm poly}(\log\widetilde{N}\log N))
\end{equation}
elementary quantum gates for each $t$. This is the time complexity of our
algorithm for handling each generation. It is quadratically faster than classically expected
amount considering the fact that the number of combinations is $O(\widetilde{N}^2)$ in our
crossover procedure (see the next subsection for the details).
As for the space complexity, we spend ${\rm poly}(\log N)$ qubits and $O(\log N)$
classical bits as is easily evaluated from the description of Algorithm \ref{ouralg} (Fig.~\ref{figouralg}).

\subsection{Comparison with a classical counterpart}\label{seccomparison}
The classical counterpart of our algorithm is the one described in Fig.~\ref{figclassicalalg}
(Algorithm \ref{classicalalg}). As is obvious
from its structure, it should have the same output and the same number of iterations
as our algorithm with almost certainty as long as the same random seed is used for
step 1 to construct $R$ for each value of $t$.
\begin{figure*}[h]
\begin{algorithm}{}\label{classicalalg}
\hspace{-1.5mm}Consider a threshold $f_{\rm th}$ for fitness values, which is considered to be
sufficiently large.\\
$t\longleftarrow 0$.\\
{\bf REPEAT} 1.-7.:
\begin{itemize}
\item[1.] Construct the pseudo-randomizer $R$ found in Algorithm \ref{ouralg} (Fig.~\ref{figouralg})
as a classical function. Fix $R$ until next $t$. Using $R$, we generate a generation $X=\{R(q)\}$ with
$q=0_0\cdots0_{c-2}1_{c-1},\ldots,1_0\cdots1_{c-1}$. We require
$\widetilde{N}=2^c \lnapprox N=2^n$.\\
(Thus, we have $\widetilde{N}-1$ $n$-bit-length chromosomes in $X$ presently.)
\item[2.] {\bf IF} $t\not=0$ {\bf THEN} put $z$ into $X$
($z$ is the best chromosome found in the $(t-1)$th trial) {\bf ELSE}
generate a random $c$-bit string $\gamma$ and put $R(\gamma)$ into $X$ {\bf ENDIF}\\
(Now we have $\widetilde{N}$ chromosomes in $X$.)
\item[3.] Split all $x\in X$ between the $(l-1)$th bit and the $l$th bit.
This makes $l$-bit strings $x^{\rm left}$'s and $(n-l)$-bit strings $x^{\rm right}$'s.
Generate the set $\hat X$ consisting of all $x$'s and all of their children that are
all the combinations of $x^{\rm left}$'s and $x^{\rm right}$'s.
\item[4.] Apply the template-based mutation \verb|tmp_mut|, introduced in Sect.~\ref{secmutation},
as a classical procedure to all chromosomes in $\hat X$.
\item[5.] Find the chromosome $z$ having the best fitness value among those in $\hat X$.
\item[6.] {\bf IF} $f(z)\ge f_{\rm th}$ {\bf THEN} {\bf RETURN} $z$ and {\bf EXIT}~~{\bf ENDIF}
\item[7.] $t\longleftarrow t+1$.
\end{itemize}
\end{algorithm}
\caption{\label{figclassicalalg}Description of a classical counterpart of our algorithm.}
\end{figure*}
The computational costs of individual procedures in Algorithm\ \ref{classicalalg} are
as follows.
\begin{itemize}
\item
Generating $X$ in the steps 1.-2. takes $O({\widetilde{N}}{\rm poly}(cn))$
basic operations.
\item
We need to use $O(n{\widetilde{N}}^2)$ space and $O(n{\widetilde{N}}^2)$
basic operations to perform all the crossovers among $\widetilde{N}$ parents.
\item
Mutations acting on the individuals of $\hat X$ take $O(n{\widetilde{N}}^2)$ basic operations.
\item
The selection to find the best individual from $\hat X$ takes $O({\widetilde{N}}^2{\rm poly}(n))$
basic operations considering the cost ${\rm poly}(n)$ of calculating a fitness value.
\end{itemize}
As $n=\log_2 N$ and $c=\log_2 \widetilde{N}$, the time and space complexities are
$O({\widetilde{N}}^2{\rm poly}(\log N))$ and $O({\widetilde{N}}^2\log N)$, respectively,
for each $t$, i.e., for handling each generation.

In contrast, as we have seen in Sect.~\ref{sectotalcost}, the time and space
complexities of our quantum genetic algorithm (described in Fig.~\ref{figouralg})
are $O({\widetilde{N}}{\rm poly}(\log\widetilde{N}\log N))$ and ${\rm poly}(\log N)$, respectively,
for each $t$. Therefore, neglecting the difference between ${\rm poly}(\log \widetilde{N}\log N)$
and ${\rm poly}(\log N)$, we have achieved
a quadratic speedup over its classical counterpart together with an exponential reduction
in space.

The classical counterpart has been constructed by keeping the one-by-one correspondence with
the quantum algorithm. Thus there is a possibility that a better classical algorithm with
the same behavior exists. This is in fact the case for Algorithm~\ref{classicalalg}.
For a fairer comparison, now we reform Algorithm~\ref{classicalalg} and reduce its space
complexity. The algorithm described in Fig.~\ref{figclassicalalg2} (Algorithm \ref{classicalalg2})
has the same output and the same number of repetitions as Algorithm~\ref{classicalalg} while its
space complexity is exponentially reduced. We use the same pseudo-randomizer construction and
the same mutation procedure as before. Of course, we set $\widetilde{N}=2^c$ and $N=2^n$ for
integers $c$ and $n$ satisfying $1\le c < n$.
\begin{figure*}[h]
\begin{algorithm}{}\label{classicalalg2}
\hspace{-1.5mm}Consider a threshold $f_{\rm th}$ for fitness values, which is considered to be
sufficiently large.\\
$t\longleftarrow 0$.\\
{\bf REPEAT} 1.-8.:
\begin{itemize}
\item[1.] Construct the pseudo-randomizer $R:\{0,1\}^c\rightarrow\{0,1\}^n$ as a classical operation.
\item[2.] Construct the mutation process as a map $M$ as a classical operation.
\item[3.] {\bf IF} $t=0$ {\bf THEN} for a random $\gamma\in\{1,\ldots,\widetilde{N}-1\}$,
          $z\longleftarrow R(\gamma)$ (otherwise, $z$ is the best chromosome found in the $(t-1)$th step) {\bf ENDIF}
\item[4.] For a random $\gamma'\in\{1,\ldots,\widetilde{N}-1\}$, $j\longleftarrow R(\gamma')$.
\item[5.] {\bf FOR} $a\longleftarrow 0$ {\bf TO} $\widetilde{N}-1$:\\
$~~~~~${\bf IF} $a=0$ {\bf THEN} $x\longleftarrow z$ {\bf ELSE} $x\longleftarrow R(a)$ {\bf ENDIF}\\
$~~~~~${\bf FOR} $b\longleftarrow 0$ {\bf TO} $\widetilde{N}-1$:\\
$~~~~~~~~~~${\bf IF} $b=0$ {\bf THEN} $y\longleftarrow z$ {\bf ELSE} $y\longleftarrow R(b)$ {\bf ENDIF}\\
$~~~~~~~~~~$Crossover $x$ and $y$ and obtain children $v$ and $w$.\\
$~~~~~~~~~~$Find the best chromosome $g$ among the chromosomes $M(x)$,\\
$~~~~~~~~~~$$M(y)$, $M(v)$ and $M(w)$.\\
$~~~~~~~~~~${\bf IF} $f(g)>f(j)$ {\bf THEN} $j\longleftarrow g$ {\bf ENDIF}\\
$~~~~~${\bf END FOR}\\
{\bf END FOR}
\item[6.] $z\longleftarrow j$.
\item[7.] {\bf IF} $f(z)\ge f_{\rm th}$ {\bf THEN} {\bf RETURN} $z$ and {\bf EXIT}~~{\bf ENDIF}
\item[8.] $t\longleftarrow t+1$.
\end{itemize}
\end{algorithm}
\caption{\label{figclassicalalg2}Description of an improved classical counterpart of our algorithm.}
\end{figure*}
In this algorithm, $R$ is called $O(\widetilde{N}^2)$ times for each value of $t$. We know
that $R$ internally takes ${\rm poly}(cn)$ time. Therefore, this algorithm spends
$O(\widetilde{N}^2{\rm poly}(\log \widetilde{N}\log N))$ time for handling each generation.
As for the space complexity, it spends only ${\rm poly}(\log N)$ space, which is clear from
the algorithm structure. 

In comparison to this enhanced classical algorithm, our quantum algorithm still
has a quadratically small time complexity as shown in Eq.~(\ref{cxtime}). 

\section{Discussion}\label{secDis}
How to perform crossovers in a quantum manner was a pending problem in conventional
quantum genetic algorithms \cite{UPV04,MBC04,UPV06,MBC08}. In fact, a selective
crossover for specific two chromosomes is expensive when they are component states
of a superposition. Even if we attach address states pointing to the component states,
we need to look up the classical data of the chromosomes to construct a quantum circuit
realizing the unitary operation for this purpose, or more specifically, for placing
address-controlled bit-flip gates appropriately. In our algorithm (Algorithm~\ref{ouralg}
shown in Fig.~\ref{figouralg}),
we have avoided to mimic a classical way and chosen a different approach.
We use two identical copies of a superposition corresponding to a generation and
utilize relabelling of qubits so as to handle all possible combinations of substrings
simultaneously. Obviously, the classical counterpart of our algorithm is the one that
seeks for the best chromosome (after a mutation) among all possible crossovers for a
chosen crossing site for each generation.
Comparing our algorithm with the classical counterpart, we concluded that we have
achieved a considerable reduction in the computational cost.

One may, however, claim that usually at most several crossovers are performed for
a single generation in a classical genetic algorithm. Indeed, our algorithm is not
aimed to be a quantum counterpart of a common classical genetic algorithm.
As we discussed in Sect.~\ref{secM}, a straightforward conversion of a common classical
algorithm into quantum one by simply incorporating a quantum search into the selection
procedure has a problem: we need to either interpret nonconsecutive integers to consecutive
ones or use an expensive construction for the inversion-about-average operation in order
to perform the Grover-BBHT search, which causes a significant loss of performance. This
problem should be resolved so as to find a meaningful quantum counterpart for the common
case. Seemingly, the following workaround looks fine: (i) Use the pseudo-randomizer $R$
used in Algorithm\ \ref{ouralg} instead of a random number generator to generate initial
chromosomes of a generation.  (ii) Apply a small number of crossovers. (iii)
Use a selection procedure similar to that of Algorithm\ \ref{ouralg}.
Nevertheless, as we have discussed, it is not known how to construct the procedure (ii)
as a unitary operation without the expensive process of looking up classical data of 
chromosomes. Therefore, it is the fact that a meaningful quantum counterpart is not easily
found for a common classical genetic algorithm.

In view of the search space covered by each generation, a simultaneous crossover is,
of course, desirable. Use of a superposition for this purpose was discussed \cite{UPV04,UPV06}
but not developed previously. In this sense, we have made a meaningful improvement
by introducing Algorithm\ \ref{ouralg}.

Besides the crossover, let us discuss on the selection procedure. Our algorithm uses
a variant of the Grover-BBHT search to achieve a quadratic speedup over its classical
counterpart. The internal cost for each query is kept polynomial in the length of a
chromosome because of the polynomial cost of our pseudo-randomizer, as described
in Sect.~\ref{seccost}. Apart from the complexity, there is some room to find a different
design for the selection. Our algorithm is designed to carry over only the best
chromosome $z$ to the next generation, among the chromosomes existing after the quantum
crossover and mutation procedures. Since projective measurements are used in the quantum
selection, it is inevitable to demolish other chromosomes. This can be a drawback
because some of them may possess good fitness values albeit not the best. To mitigate this
severe selection, one may keep the values of $z$ obtained in several elder generations as
classical data. These values can be put into a later generation by modifying the
register-initialization procedure slightly: one can modify the pseudo-randomizer so that
it does not touch several input strings; then one can map them to the kept values of $z$.
In this way, one may maintain a better diversity for high-fitness chromosomes. This is one
possible extension of our algorithm.

There have not been many studies on quantum genetic algorithms so far. It is hoped that
several or more different designs of genetic procedures will be developed for quantum
computers.


\section{Summary}\label{secCon}
We have proposed a genetic algorithm whose crossover, mutation and selection procedures
have been all constructed as quantum routines so that quantum parallelism is effectively used.
Its crossover procedure performs crossovers among all chromosomes of a generation. The run time
of our algorithm to update each generation is quadratically faster than that of its classical
counterpart, apart from negligible factors.

\begin{acknowledgements}
A.S. is thankful to Shigeru Yamashita for his comment.
A.S. and M.N. were supported by the ``Open Research Center'' Project
for Private Universities: matching fund subsidy from MEXT.
R.R. is supported by Industry Canada and CIFAR.
\end{acknowledgements}

\appendix
\section{An example of constructing the pseudo-randomizer $R$}
In this appendix, we show an example to construct the pseudo-randomizer $R$
used in Algorithms~\ref{ouralg},~\ref{classicalalg}~and~\ref{classicalalg2}.
It should map a $c$-bit string to a pseudo-random $n$-bit string except for
$0_0\cdots0_{c-1}$ that is mapped to $0_0\cdots0_{n-1}$. Its internal circuit
complexity should be ${\rm poly}(cn)$. As we use a quantum circuit to realize
it in Algorithm~\ref{ouralg}, it is desirable to employ a circuit structure
that is originally unitary.

Consider inputs $a\in\{0,1\}^c$. We design a circuit that maps $a_0\cdots a_{c-1}0_0\cdots0_{n-1}$
to $a_0\cdots a_{c-1}\kappa_0\cdots \kappa_{n-1}$ with $\kappa=R(a)$, an $n$-bit pseudo-random number
(here, $a_0\cdots a_{c-1}$ and $\kappa_0\cdots \kappa_{n-1}$ are the binary representations of $a$
and $\kappa$). By the definition of $R$, the circuit preserves $0_0\cdots0_{c-1}0_0\cdots0_{n-1}$.
This circuit is generated by function \verb|gen_r_circ|() described in Fig.~\ref{fig_gen_r_circ}.
\begin{figure*}[h]
\begin{routine}
{\bf FUNCTION} \verb|gen_r_circ|():\\
Comment: We use wires $v_0,\ldots,v_{c-1},w_0,\ldots,w_{n-1}$.\\
{\bf FOR} $i\longleftarrow 0$ {\bf TO} $c-1$:
\begin{itemize}
\item[(1)] Use a random number generator to generate an $n$-bit integer $\gamma$.
Write its binary representation as $\gamma_0\cdots\gamma_{n-1}$.
\item[(2)] Using the wire $v_i$ as the control wire (namely, the control bit), output
the gate {\em controlled-}$(X_0)^{\gamma_0}\otimes\cdots\otimes(X_{n-1})^{\gamma_{n-1}}$
with $X_k$ the bit flip gate acting on the wire $w_k$ ($k=0,\ldots,n-1$).
In this gate, the bit flips are active under the condition that $v_i=1$.
\end{itemize}
{\bf END FOR}
\end{routine}
\caption{\label{fig_gen_r_circ}Description of function {\tt gen\_r\_circ}().}
\end{figure*}
As is clear from the description, the circuit output from this function can be directly used
as a quantum circuit. Using the circuit $C=\;$\verb|gen_r_circ|(), we have
$|a\rangle|0\rangle\overset{C}{\mapsto}|a\rangle|R(a)\rangle$~~$\forall a\in\{0,1\}^c$.
The circuit complexity of $C$ is $O(cn)$ because, for each $i$, at most $n$ CNOT
gates are used to decompose the gate output from step (2). In addition, \verb|gen_r_circ|()
spends $O(c\;{\rm poly}(n))$ time when a common random number generator \cite{Knuth97,MN98}
is used in step (1).

Note that the function \verb|gen_r_circ|() is called only once for each $t$, in the
beginning of step 1 in Algorithms\ \ref{ouralg},~\ref{classicalalg}~and~\ref{classicalalg2}.
We have only to reuse the circuit $C$ for the use of the pseudo-randomizer until $t$ is
incremented.

It is expected that outputs from the circuit $C$ possess good uniformity if we use
a good random number generator in step (1) of \verb|gen_r_circ|() for generating $C$.
Let us write $\gamma$ as $\gamma(i)$ to emphasize its dependence on $i$.
For a nonzero input $a_0\cdots a_{c-1}$, the $k$th bit of the output 
$R(a)$ is $\sum_{i=0}^{c-1}a_i\cdot\gamma_k(i)~~{\rm mod}~2$. This indicates
that, for two different inputs $a$ and $a'$, the $k$th bits of $R(a)$ and $R(a')$
differ with probability $1/2$ in the ideal case where $\gamma(i)$'s are generated
from a true random number generator. This is because $a$ and $a'$ differ by
at least a single bit. It also indicates that two different bits, the $k$th and the
$k'$th bits, of $R(a)$ for a nonzero input $a$ differ with the probability $1/2$ in
the ideal case. This is because $\gamma_k(i)$ and $\gamma_{k'}(i)$ differ with the
probability $1/2$.

Now we show the result of our numerical test of $C$. We tried statistical
tests of randomness \cite{Knuth97,NIST2001} to test pseudo-random numbers output
from $C$, using NIST's Statistical Test Suite (STS) (version 2.1.1) \cite{NIST2001}.
We set $c=10$ and $n=32$. Mersenne Twister (MT) (version mt19937ar) \cite{MN98} was
used to generate $\gamma$ in step (1) of \verb|gen_r_circ|(). We used the seed value
$121212$ and did not reset MT during the circuit generation. The circuit $C$ output
from \verb|gen_r_circ|(), of course, consisted of $10$ outputs from step (2). For
this $C$, we used the inputs $a\in\{0,1\}^c\backslash\{0_0\cdots0_{c-1}\}$ from
smaller to larger and obtained corresponding outputs $R(a)$ by numerical computation.
We obtained $1023\times 32$ bits in total in the outputs, since $2^{10}-1=1023$. We
regarded them as a serial bit string from left to right and used STS in its default
setting to test the string. In the execution of STS, we used $25$ binary sequences with
length $1200$ as samples from the string.
The following tests were tried with the default parameter values in STS: the Frequency
Test, the Block Frequency Test, the Cumulative Sums Test, the Runs Test, the Longest-Run-of-Ones
Test, the Binary Matrix Rank Test, the Spectral DFT Test and the Serial Test. The string
passed the tests except for the Binary Matrix Rank Test. It should be noted that the input
length was too small for the binary matrix rank test \cite{NIST2001}. In addition, randomness
is not very strictly required for the use in evolutionary computing. Therefore, considering the
tests that the string passed, we may claim that \verb|gen_r_circ|() generates a usable
pseudo-randomizer circuit for our algorithm.

We conducted another test: We generated ten circuits by calling \verb|gen_r_circ|()
ten times without resetting MT, using the seed value $676767$. For each circuit, we
performed the same process as above to obtain the serial bit string. We obtained ten
serial bit strings in total and tested the concatenated string using STS. As samples
input to STS, we used $25$ binary sequences with length $12000$. The concatenated string
passed the tests except for the Binary Matrix Rank Test and the Spectral DFT Test.
It was unexpected that it did not pass the spectral DFT test. It requires a further
investigation to reveal the reason of this phenomenon.

The results of the first and the second tests are summarized in Table\ \ref{ttest}.
In summary for this appendix, we found that a pseudo-randomizer circuit whose outputs
possess enough randomness for the use in evolutionary computing can be generated by the
function \verb|gen_r_circ|(). It is hoped that the function will be improved so as to
achieve better randomness for the sake of general use.

\begin{table*}[tbp]\begin{center}
\caption{\label{ttest}List of test results for the first and the second tests
we performed (see the text for the details of the tests).
Each value is a $P$-value (see Ref.~\cite{NIST2001} for its meaning for
each test). The asterisk $*$ indicates a failure to pass the test.
We should mention that the values for the Longest-Run-of-Ones Test accidentally
coincided, while the internally-used results for sample sequences were different.}
\begin{tabular}{|r|l|l|}
\hline
&$~$1st Test&$~$2nd Test\\
\hline
Frequency Test&0.021262&0.186566\\
Block Frequency Test&0.105618&0.001156\\
Cumulative Sums Test (Forward)&0.105618&0.029796\\
Cumulative Sums Test (Reverse)&0.001691&0.057146\\
Runs Test&0.141256&0.010606\\
Longest-Run-of-Ones Test&0.875539&0.875539\\
Binary Matrix Rank Test&0.000000~$*$&0.000000~$*$\\
Spectral DFT Test&0.000533&0.000000~$*$\\
Serial Test&0.041438&0.186566\\
\hline
\end{tabular}
\end{center}
\end{table*}

\bibliographystyle{spmpsci-mod}
\bibliography{refs_search_qip}

\end{document}